# Feature Attention Network (FA-Net): A Deep-Learning Based Approach for Underwater Single Image Enhancement

Muhammad Hamza[a], Ammar Hawbani*[a], Sami Ul Rehman[a], Xingfu Wang[a], Liang Zhao[b]
[a]Computer Science and Technology, University of Science and Technology of China, JinZhai, Hefei, 230026, Anhui, China.; School of Computer Science, Shenyang Aerospace University, Daoyi South, Shenyang, 110136, Liaoning, China.

## ABSTRACT

Underwater image processing and analysis have been a hotspot of study in recent years, as more emphasis has been focused to underwater monitoring and usage of marine resources. Compared with the open environment, underwater image encountered with more complicated conditions such as light abortion, scattering, turbulence, nonuniform illumination and color diffusion. Although considerable advances and enhancement techniques achieved in resolving these issues, they treat low-frequency information equally across the entire channel, which results in limiting the network's representativeness. We propose a deep learning and feature-attention-based end-to-end network (FA-Net) to solve this problem. In particular, we propose a Residual Feature Attention Block (RFAB), containing the channel attention, pixel attention, and residual learning mechanism with long and short skip connections. RFAB allows the network to focus on learning high-frequency information while skipping low-frequency information on multi-hop connections. The channel and pixel attention mechanism considers each channel's different features and the uneven distribution of haze over different pixels in the image. The experimental results shows that the FA-Net propose by us provides higher accuracy, quantitatively and qualitatively and superiority to previous state-of-the-art methods.

## 1. INTRODUCTION

In recent years, academics have been more interested in examining the mysterious underwater environment. With the rapid advancements in marine investigation and observation, there is a growing interest in researching this enigmatic part of the ocean. Due to the harsh underwater environment and unpleasant lighting conditions, examining the images is a highly complex task. Underwater images often deteriorate due to wavelength absorption and forward and backward scattering[1]. In addition, aquatic snow creates noise and increases dispersal effects[2]. These detrimental effects reduce visibility, contrast, and even produce color variations, causing in limiting the practical use of underwater images in marine biology, aquatic. ecosystems, archaeology[3,4].

Considering the light decay, haze effect and ill-posedness, numerous methods have been proposed to improve underwater images using the dark channel prior[5], the light channel prior[6], the maximum intensity prior[7], and the blurriness prior[8]. Dark channel prior (DCP)[5] is a well-known prior method. The haze-free image patches often have low intensity values at the last channel based on the assumption of the dark channel prior. However, if prior assumptions are wrong[9], prior-based methods can lead to significant errors in estimation and can produce artifacts. Rather than adopting the prior-based approaches, deep learning-based methods are being used in many computer vision problems[10], including super-resolution enhancement (SESR)[11], Adversarial Networks (GAN)[12], and Underwater Image Enhancement Benchmark (WaterNet)[13]. These approaches use deep learning models trained on synthetic data sets. Compared to prior-based methods, deep learning techniques attempt to improve the final image directly.

Considering the robustness and performance improvement using deep learning, in this work, a CNN (FA-Net) model is developed to enhance underwater images. Despite recent efforts, deep learning methods remain an issue because the contrast in images is often unpredictably distributed. Attention mechanism[14,15] is widely used in neural networks and plays an important role in network performance. The advent of ResNet[16] made it possible to create deep networks. Inspired by the work[15,16,17], we propose the RFA (residual feature attention) module. The RFA module provides the flexibility for handling different types of information. by combines channel attention and pixel attention mechanisms to handle different features and pixels differently. We adopt the notions of skip connection[16] and attentional mechanisms to develop a building block that consists of skip connections and local residues learning. In general, feature fusion network (FA-Net) is based on channel attention [17], deep residual learning[16] and pixel attention[18].

## 2. RELATED WORK

In recent years, much attention has been paid to the restoration and enhancement of underwater images to encourage the study of the marine environment. Recently, several techniques have been proposed based on underwater image formation model, including color correction, and enhancing of underwater images, to combat poor visibility and color cast in underwater images. Therefore, image enhancement methods are always more in demand than other methods. According to Revised Underwater Image Formation Model[19], the physical model of the underwater image is represented as:

$$I_c = J_c e^{-\beta_c^D(v_D) \cdot z} + B_c^\infty \left(1 - e^{-\beta_c^B(v_B) \cdot z}\right) \quad (1)$$

Where $I_c$ is the captured image with $c$ RGB color channel. $J_c$ is the image with the unattenuated signal, $B_c^D$ is denoted as wideband direct signal attenuation coefficient and $B_c^B$ is the backscatter wideband attenuation coefficient, $B_c^\infty$ represents the veiling light for the water type, $V_D$ and $V_B$ denotes the coefficient dependencies with respect to direct signal and backscatter and finally $z$ denotes the image range along line-of-sight.

Dark Channel Prior (DCP) is a widely used algorithm for image enhancement based on underwater image formation model. DCP was initially thought to extract an external image by assuming that the brightness of an object at least one-color channel is close to zero. For example, in Wavelength-Compensation[3], haze and color variations are removed by combining the DCP with a wavelength-dependent compensation algorithm. Blurriness prior depth estimation[8] method based on image blur and light absorption is proposed for image dehazing. A combined method of a priori adaptive attenuation curve and underwater light propagation characteristics[20] is proposed to enhance an underwater image. While these methods have had high successes ratio, methods based on prior are not robust enough to handle all situations, such as an unrestricted natural environment.

With the accomplishment of deep learning in image processing tasks and the accessibility of substantial image datasets, numerous models have been proposed based on convolutional neural networks. Recently, WaterGAN a semi-supervised deep learning-based model for underwater image enhancement network has been proposed[21]. In the first step, WaterGAN considers an underwater attenuation, scattering, and vignetting generation model to simulate imagery and depth estimation. In the second step, WaterGAN uses CNN for color correction. CycleGAN based Generative Adversarial Network (UGAN)[22] is proposed to improve underwater images. Taking advantage of the adversarial network, it allows the underwater images to be taken in unknown places using the unique images generated by CycleGAN[22]. The introduction of skip connection In ResNet[16] allowed the network to learn deeper feature representations. DenseNet[23] substantially reduced the number of parameters by strengthening feature propagation and reusing the features while attaining impressive results. RCAN[17] introduces the channel attention (CA) mechanism. RCAN adaptively rescale each channel feature by modeling the interdependencies across each feature channel. Similarly, our FA-Net comprises the channel and pixel attention mechanism, combined with deep residual learning and long-short skip connection.

**Contributions:**

Overall, our contributions in this paper are as below:

- We offer a deep learning-based feature attention framework for single image dehazing based on enhancing characteristics for underwater images. FA-Net is an end-to-end neural network. We present a residual feature attention module (RFA) combined with a channel and pixel attention. This module provides flexibility in handling different types of information, focusing more on thick haze pixels and more channel information.
- We propose a group structure that consists of a residual feature attention module. The long skip connection in group structure passes the feature information from the shallow to the deeper in the network. It helps to learn more desirable information by bypassing the unrelated low-frequency information.
- We conduct numerous experiments to prove our FA-Net. FA-Net is far superior to the previous underwater image-enhancing methods, especially on the hazy tick texture, and achieves far more accurate results than previous state-of-the-art methods on multiple datasets.
- Additionally, we perform low-level vision tasks, image depth estimation, and salience object detection on enhanced images and obtain better result.

We have organized our article as follows. Section 3 contains the proposed network with FA module and loss function and analysis of our network. Section 4 describes the experimental results, performance compared to conventional methods and ablation studies. We also demonstrate that our model is efficient for lol-level vision task. Section 5 is the conclusion.

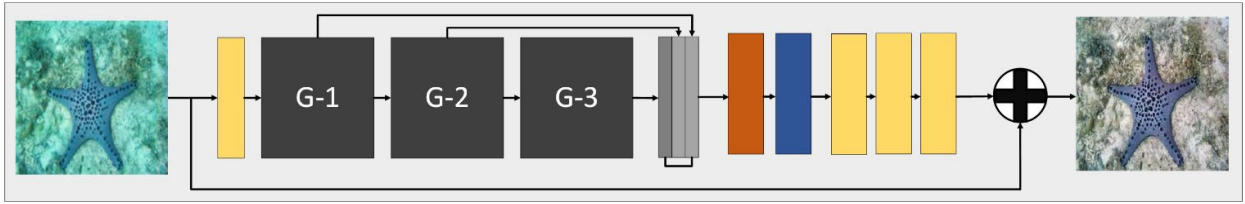

Fig.1. The overall FA-Net architecture. Black blocks are Group, the blue block is pixel, and Sienna block is channel attention

## 3. PROPOSED METHOD

Our FA-Net is a fusion structure subdivided into two main blocks: the RFA block with residual learning and the group structure to perform the fusion mechanism and add depth to the network. The FA-Net input is a hazy image, as shown in fig.3, and passes through the N group block architecture. The group module contains the residual feature attention module combined with the local residual learning, pixel attention and channel attention. Finally, the features fetched from the RFAB and group module, moves to the reconstruction part and thus obtain the enhanced image, as shown in fig.1.

### 3.1 Residual Feature Attention Block

The residual feature attention block (RFAB) structure includes pixel and channel attention modules combined with a local residual training learning. Local residual learning allows fine haze or low-frequency information to bypass the shallow to the deep part of the network through numerous local residual connections. The focus is the efficient information transfer, extract different pixels and channels-wise features of distinct weights to improve network performance and training stabilization. It consists of channel attention, pixel attention unit, and local residual learning, as shown in Fig.2.

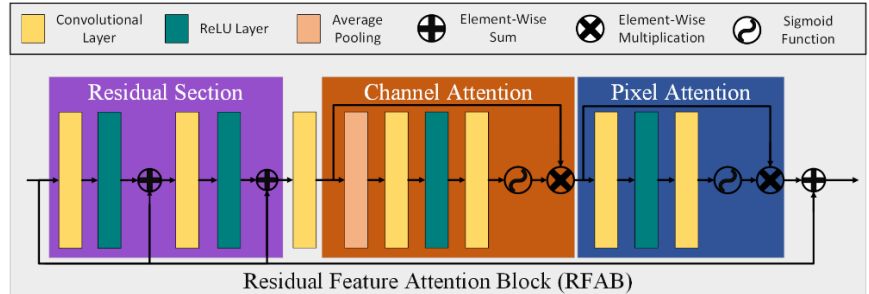

Fig.2. The proposed Residual Feature Attention Block. Purple section is Residual Learning, Sienna section is Channel Attention, and the Blue section is Pixel Attention

There are two residual training portions, and each portion consists of a 3×3 convolutional layer followed by a rectified linear block (ReLU) and ⊕ denotes the element-wise sum. The Channel Attention module is to generate different channel weights in a channel-wise dimension. It is essential for the channel attention module to provides efficient and accurate channel-wise weighting. Using the global averages pooling, we take the global spatial information for each channel in the channel descriptor.

$$G_c = R_p(Z_c) \qquad (2)$$

Where $R_p(Z_c)$ is the global average pooling. After applying the GAP, the feature map changes the shape from C×H×W to C×1×1. To get the weights of the different channels, features lead through two convolution layers and sigmoid, ReLU activation function later.

$$F_{ca} = \sigma\Big(Conv\big(\delta(Conv(G_c))\big)\Big) \qquad (3)$$

Whereas δ represents ReLU and σ denotes the sigmoid function. Ultimately, we employ element-wise multiplication to input and Pixel Attention module.

$$F_c = F_{ca} \otimes Z_c \tag{4}$$

Considering the uneven distribution of haze, the pixel attention module forces the network to pay more attention to informative features such as haze pixels and the high-frequency region of the image. Same as channel attention we feed the channel attention input value directly into the two convolutional layers using ReLU and the sigmoid activation function, as shown in Equ.5.

$$Z_p = \sigma\left(Conv\left(\delta(Conv(I_c))\right)\right) \tag{5}$$

Whereas δ represents ReLU, σ denotes the sigmoid function, and $I_c$ is the input. Finally, we perform element-wise multiplication to Pixel Attention and input.

$$F_p = I_c \otimes Z_p \tag{6}$$

Ultimately, the final equation, combining the two residual learning section, channel attention and pixel attention can be written as:

$$R\ F\ A = \oplus\left(\left(\delta(Conv(I))\right) \oplus \left(\delta(Conv(I))\right) \oplus (Conv(I), F_c, F_p)\right) \tag{7}$$

### 3.2 Group Structure

The group architecture consists of residual feature attention block structures which are sequentially connected followed by convolutional layer and with skip connection. The long-term skip connection helps the model to pass the rich information through the layers and the other group block. Perpetual residual feature attention blocks add depth and expression to the FA-Net and help in contribution of multi-level information fusion. Fig 3. illustrates the group structure.

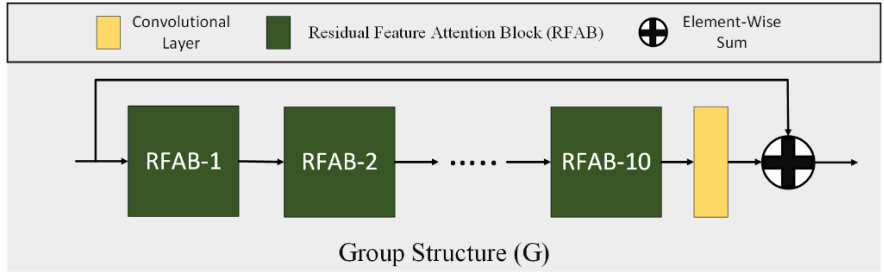

Fig.3. Group Structure (G).

### 3.3 Network Architecture

FA network is based on FFA-Net and U-Net that is wildly used for image dehazing and image enhancement. As shown in fig.3. Two residual part is connected with channel attention and pixel attention for the generation of our RFA block, which aims at learning map the different feature maps. Same as FFA-Net we use the 3-Group Block, and every group block contains the sequentially connected B=10 residual feature attention blocks (RFAB) and a convolutional layer. We set the kernel size to 1×1 of the convolution layer in the channel attention module. Except for the channel attention, we set the kernel size to 3×3 in all the other convolutional layers. Every group and RFAB output are 16 filters to keep the size fixed. Finally, we add a two-layer convolutional network and a residual learning module at the end of the network. In the end, we restored the required enhanced image.

### 3.4 Loss Function

We trained our FA-Net on SSIM loss and L1 (MAE) loss to get superior results, and these also to help improve the learning process. DRN[15] proves that L1 loss achieves superior results than L2 loss in terms of PSRN and SSIM. So, we adopt the L1 or mean absolute error loss and can be expressed as:

$$L_1 = \frac{1}{N}\sum_{i=1}^{N}|I_{gt}^i - FA(I_{haze}^i)| \tag{8}$$

Whereas $I_{gt}^i$ is the ground truth input, $I_{haze}^i$ is the hazy image input. We also use SSIM loss to measure the structural similarity between ground truth and hazy images. SSIM is a perceptual loss that considers image degradation in terms of structural information between two images based on two vital perceptual phenomena: brightness and contrast. Because of these phenomena, we use SSIM loss for better training and to improve enhanced image structure and can be computed as:

$$\text{SSIM}(I_{gt}, I_{hz}) = \frac{(2\mu_{I_{gt}}\mu_{I_{hz}} + c_1)(2\sigma_{I_{gt}I_{hz}} + c_2)}{(\mu_{I_{gt}}^2 + \mu_{I_{hz}}^2 + c_1)(\sigma_{I_{gt}}^2 + \sigma_{I_{hz}}^2 + c_2)} \quad (9)$$

Where $I_{gt}, I_{hz}$ is the ground truth and hazy image, respectively. $\mu_{I_{gt}}, \mu_{I_{hz}}$ are the average values, $\sigma_{I_{gt}}, \sigma_{I_{hz}}$ are the variance and $\sigma_{I_{gt}I_{hz}}$ is the covariance of $I_{gt} and I_{hz}$ , respectively? C1 and C2 are the two variables to maintain stability. As the rage of SSIM is from 0 to 1, the SSIM can be expressed as follow:

$$L_t = L_1 + L_s \quad (10)$$

## 4. EXPERIMENTS AND RESULTS

### 4.1 Datasets, Metrics and Training Settings

We adopt the UFO-120[11] and EUVP[24] dataset for evaluation in this work. UFO-120 contains the 1500 raw hazy images for training and validation with corresponding high-resolution references images and another 120 images for testing. Furthermore, the EUVP dataset contains 12K training samples and 515 images for evaluation and validation. We adopt the model trained on UFO-120 for the EUVP. We conduct a quantitative analysis by comparing the predicted results with the reference haze free images in terms of MAE, PSNR, and SSIM. The batch of pared, ground truth and hazy images with the size of 256×256 passed to FA-Net as input to speed up the training and satisfactory results. FA-Net is trained for 500 steps. We use the Adam optimizer with default values of $\lambda_1 = 0.9$ and $\lambda_2 = 0.999$, respectively. We set the initial learning rate to 0.0005 and reduced the to 0.00002 after every 20 epochs. Tensor-flow is used for the implementation on Intel(R) Core (TM) i7-9700 CPU @ 3.00GHz with NVIDIA RTX 2080 SUPPER GPU and 16 GB of RAM.

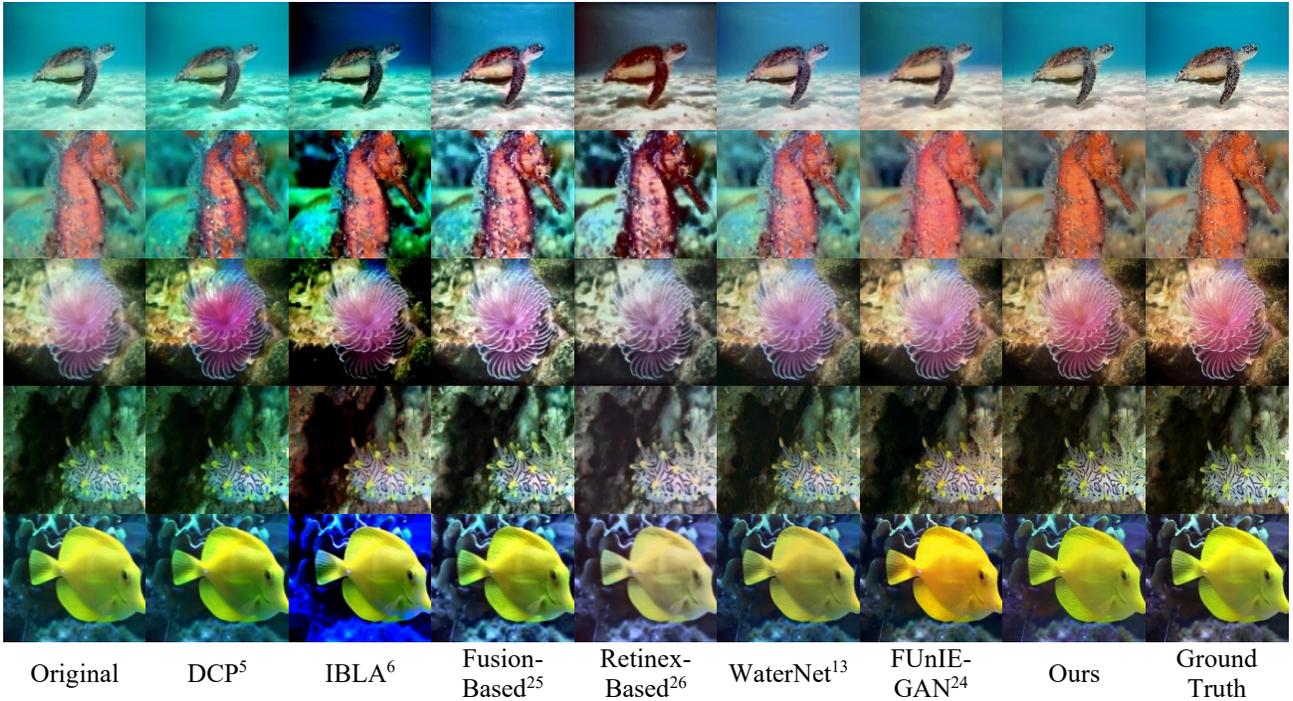

Original | DCP[5] | IBLA[6] | Fusion-Based[25] | Retinex-Based[26] | WaterNet[13] | FUnIE-GAN[24] | Ours | Ground Truth

Fig.4. Qualitative comparisons for SOTS on UFO-120 and EUVP dataset

### 4.2 Results Analysis and Comparison

In this section, we compare our FA-Net, quantitatively and qualitatively with previous state-of-the-art underwater image enhancement techniques. We compare our method with 6 state-of-the-art methods: DCP[5], IBLA[8], WaterNet[13], FUnIE-GAN[24], Fusion-based[25] and Retinex-based[26]. We used Mean-Absolute-Error (MAE), Structural Similarity (SSIM), Peak Signal-to-Noise Ratio (PSNR), and underwater image quality measure (UIQM)[27] for performance evaluation. The average values of MEA, PSNR, and SSIM shown in Table 1. Our proposed method is superior compared with other methods in terms of MEA, PSNR, and SSIM. In addition, we propose a comparison of the enhanced images shown in Fig. (4) for a qualitative evaluation. The improvement results evaluated by Fashion-based[25] and Retinex-based[26] are insufficient in correcting hue, color distortion, and artifacts. The methods based on the prior, DCP[5] and IBLA[8] are also insufficient in color retention and water removal. Compared with state-of-the-art methods, our proposed method is superior and provides the best performance in handling artifacts, distortion, and water removal. In addition, we calculated the Underwater Image Quality Score (UIQM)[27] for the qualitative assessment of underwater images. UIQM is based on sharpness, color, contrast, and a linear combination of UISM, UICM, and UIConM. UISM is a measure of the sharpness, UICM is a measure of the coloration, and UIConM is a measure of the contrast of an underwater image. UIQM can be represented as:

$$UIQM = c_1 \times UICM + c_2 \times UISM + c_3 \times UIConM \qquad (11)$$

where $c_1, c_2$ and $c_3$ are the scale factor, and we set the same value as those proposed in original paper [27]: $c_1 = 0.0282, c_2 = 0.2953$ and $c_3 = 3.5753$. Table 2 summarizes the average values of these metrics for UFO-120 and EUVP underwater test images used during evaluation.

Table 1: The average MAE, PSNR, AND SSIM values of enhanced results on UFO-120 and EUVP datasets

|  | UFO-120 | | | EUVP | | |
| --- | --- | --- | --- | --- | --- | --- |
|  | MAE | PSNR | SSIM | MAE | PSNR | SSIM |
| DCP [5] | 0.9108 | 18.9156 | 0.7058 | 0.9099 | 18.8853 | 0.7566 |
| IBLA [6] | 0.8765 | 17.2637 | 0.6221 | 0.9006 | 18.9590 | 0.7117 |
| WaterNet [13] | 0.9208 | 20.3775 | 0.7292 | 0.9262 | 21.0627 | 0.7921 |
| FUnIE-GAN [24] | 0.9299 | 21.4404 | 2.6952 | 0.9318 | 21.6934 | 0.7984 |
| Fusion-based [25] | 0.8825 | 16.6938 | 0.6592 | 0.8858 | 16.9699 | 0.6990 |
| Retinex-based [26] | 0.8491 | 15.1057 | 0.5869 | 0.8636 | 15.9028 | 0.6562 |
| Ours | 0.9630 | 26.4564 | 0.8487 | 0.9551 | 25.9217 | 0.8435 |

Table 2: Average UIQM results on enhanced images

|  | UFO-120 | EUVP |
| --- | --- | --- |
|  | UIQM | UIQM |
| DCP [5] | 2.110 | 3.095 |
| IBLA [6] | 2.183 | 4.280 |
| WaterNet [13] | 3.334 | 4.330 |
| FUnIE-GAN [24] | 4.749 | 4.806 |
| Fusion-based [25] | 3.448 | 4.161 |
| Retinex-based [26] | 4.220 | 4.970 |
| Ours | 5.351 | 5.356 |

## 4.3 Ablation Study

We investigated the channel attention, pixel attention and residual learning advantage to show the superior FA-Net design. Table 3 shows the results for the UFO-120 dataset. First, we used the channel attention by keeping the group structure, which retains the performance of 24.6067 in PSNR and 0.8124 in SSIM, when adopting the MAE and SSIM loss. Second, we combined the channel attention with pixel attention results in PSNR of 25.6499 and SSIM of 0.8242. Furthermore, performance increased to 26.4564 in PSNR and 0.8487 in SSIM by using the Residual Feature Attention Block (RFAB) in complete model. This demonstrates the effectiveness of residual learning in our model.

Table 3: The ablation study on UFO-120 test set for different configurations

| Channel Attention | Pixel Attention | Residual Learning | PSNR | SSIM |
| --- | --- | --- | --- | --- |
| ✓ |  |  | 24.6067 | 0.8124 |
| ✓ | ✓ |  | 25.6499 | 0.8242 |
| ✓ | ✓ | ✓ | 26.4564 | 0.8487 |

We investigated the high-level vision issue, salience object detection, and single-image depth estimation for further research. The visual results of salience object detection using BASNet[28] and depth estimation with geometric constraints[29] are shown in Fig.5. The depth of the original photographs cannot distinguish the fish from the bottom, while the depth is virtually evident on the backdrop and the fish in the predicted images. Furthermore, when compared to the same approach, the performance of significance detection on improved photos shows a substantial increase.

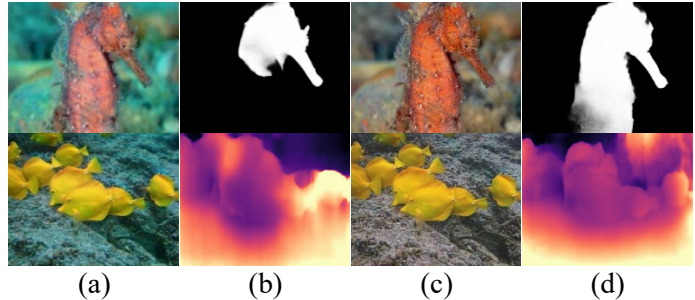

(a)  (b)  (c)  (d)

Figure.5. (a). Original images. (b). Salience object detection and single depth image result. (c). Enhanced images. (d). Salience object detection and single depth image result

## 5. CONCLUSION

Inspired by the feature attention mechanism and deep learning, in this paper a feature attention network (FA-Net) is proposed for underwater image enhancement. FA-Net consists of residual feature attention block, which consists of channel and pixel attention for efficient information transfer. Local residual learning allows fine haze or low-frequency information to bypass the shallow to the deep part of the network through several local residual connections. channel attention, pixel attention modules allow the network to generate different channel weights in a channel-wise dimension and forces informative features such as haze pixels and the high-frequency region. Comprehensive experiments with numerous datasets show that our FA-Net is superior to other state-of-the-art methods. In addition, low-level vision problems include single-image depth estimation and salience object detection expected to solve.